\documentclass[a4paper]{article}

\usepackage[pages=all, color=black, position={current page.south}, placement=bottom, scale=1, opacity=1, vshift=5mm]{background}

\usepackage[margin=1in]{geometry} 

\usepackage{amsmath,amsfonts,bm}

\def\eqref#1{equation~\ref{#1}}

\def\1{\bm{1}}

\DeclareMathAlphabet{\mathsfit}{\encodingdefault}{\sfdefault}{m}{sl}
\SetMathAlphabet{\mathsfit}{bold}{\encodingdefault}{\sfdefault}{bx}{n}

\usepackage{hyperref}
\usepackage{url}

\usepackage{mathtools}
\usepackage{amsthm}
\usepackage[numbers]{natbib}

\usepackage{times}
\usepackage{epsfig}
\usepackage{graphicx}
\usepackage{amsmath}
\usepackage{amssymb}
\usepackage{multirow}
\usepackage{caption}
\usepackage{subcaption}

\newcommand{\minisection}[1]{\vspace{0.00in} \noindent {\bf #1}\ \ }

\title{Planckian Jitter: countering the color-crippling effects of color jitter on self-supervised training}

\author{Simone Zini$^1$\thanks{Corresponding author} \and Alex Gomez-Villa$^2$ \and Marco Buzzelli$^1$  \and Bartłomiej Twardowski$^2$ \and Andrew D. Bagdanov$^3$ \and Joost van de Weijer$^2$}

\date{
	$^1$Department of Informatics, Systems and Communication, University of Milano -- Bicocca.\\ \texttt{\{simone.zini, marco.buzzelli\}@unimib.it}\\%
	$^2$Computer Vision Center, Universitat Autònoma de Barcelona \\ \texttt{\{agomezvi, btwardowski\}@cvc.uab.cat, joost@cvc.uab.es}\\
	$^3$University of Florence\\ \texttt{andrew.bagdanov@unifi.it}\\
	[2ex]%
}

\begin{document}

\maketitle

\begin{abstract}
Several recent works on self-supervised learning are trained by mapping different augmentations of the same image to the same feature representation. The data augmentations used are of crucial importance to the quality of learned feature representations. In this paper, we analyze how the color jitter traditionally used in data augmentation negatively impacts the quality of the color features in learned feature representations. To address this problem, we propose a more realistic, physics-based color data augmentation -- which we call \emph{Planckian Jitter} -- that creates realistic variations in chromaticity and produces a model robust to illumination changes that can be commonly observed in real life, while maintaining the ability to discriminate image content based on color information.
Experiments confirm that such a representation is complementary to the representations learned with the currently-used color jitter augmentation and that a simple concatenation leads to significant performance gains on a wide range of downstream datasets. 
In addition, we present a color sensitivity analysis that documents the impact of different training methods on model neurons and shows that the performance of the learned features is robust with respect to illuminant variations.
Official code available at: \url{https://github.com/TheZino/PlanckianJitter}
\end{abstract}

\section{Introduction}

Self-supervised learning enables the learning of representations without the need for labeled data~\citep{doersch2015_pretext_patches,dosovitskiy2014_pretext_surrogate}. Several recent works learn representations that are invariant with respect to a set of data augmentations and have obtained spectacular results~\citep{NEURIPS2020BYOL,chen2021SimSiam, caron2020SwAV}, significantly narrowing the gap with supervised learned representations. These works vary in their architectures, learning objectives, and optimization strategies, however they are similar in applying a common set of data augmentations to generate different image views. These algorithms, while learning to map these different views to the same latent representation, learn rich semantic representations for visual data. The set of transformations (data augmentations) used induces invariances that characterize the learned visual representation.

Before deep learning revolutionized the way visual representations are learned, features were handcrafted to represent various properties, leading to research on shape~\citep{lowe2004distinctive}, texture~\citep{manjunath1996texture}, and color features~\citep{finlayson2001solving,geusebroek2001color}. Color features were typically designed to be invariant to a set of scene-accidental events such as shadows, shading, and illuminant and viewpoint changes. With the rise of deep learning, feature representations that simultaneously exploit color, shape, and texture are learned implicitly and the invariances are a byproduct of end-to-end training~\citep{krizhevsky2009learning}.
Current approaches to self-supervision learn a set of invariances implicitly related to the applied data augmentations.

\begin{figure*}[t]
    \centering
    \includegraphics[width=\textwidth]{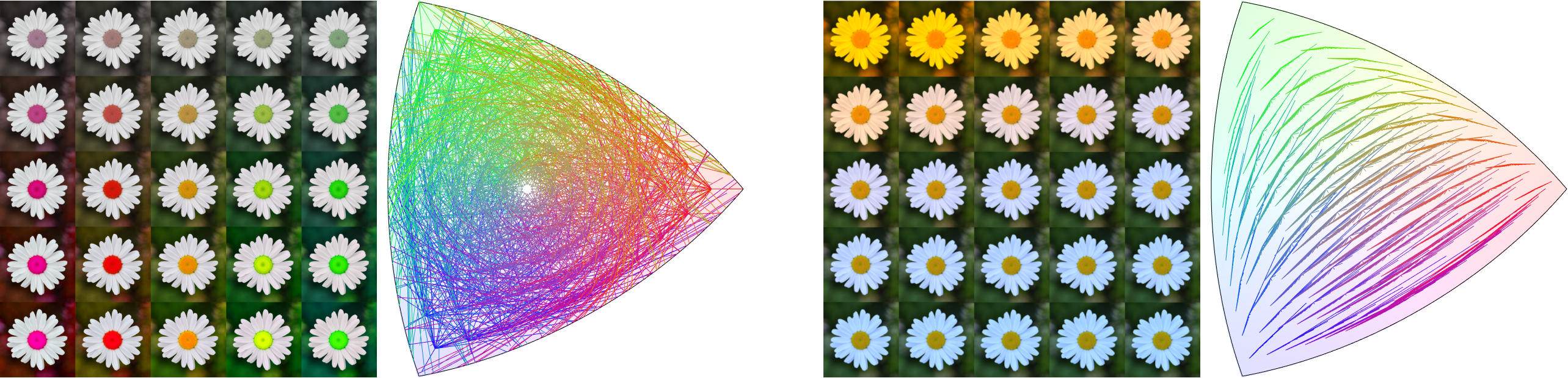}
    \caption{Default color jitter (left) and Planckian Jitter (right). Augmentations based on default color jitter lead to unrealistic images, while Planckian Jitter leads to a set of realistic ones. The ARC chromaticity diagrams for each type of jitter are computed by sampling initial RGB values and mapping them into the range of possible outputs given by each augmentation. These diagrams show that Planckian Jitter transforms colors along chromaticity lines occurring in nature when changing the illuminant, whereas default color jitter transfers colors throughout the whole chromaticity plane.
    }
    \label{fig:das_visualization}
\end{figure*}

In this work, we focus on the currently de facto choice for color augmentations. We argue that they seriously cripple the color quality of learned representations and we propose an alternative, physics-based color augmentation.
Figure~\ref{fig:das_visualization} (left) illustrates the currently used color augmentation on a sample image. It is clear that the applied color transformation significantly alters the colors of the original image, both in terms of hue and saturation. This augmentation results in a representation that is invariant with respect to surface reflectance -- an invariance beneficial for recognizing classes whose surface reflectance varies significantly, for example many man-made objects such as cars and chairs. However, such invariance is expected to hurt performance on downstream tasks for which color is an important feature, like natural classes such as birds or food. One of the justifications is that without large color changes, mapping images to the same latent representation can be purely done based on color and no complex shape or texture features are learned.
However, as a result the quality of the color representation learned with such algorithms is inferior and important information on surface reflectance might be absent.
Additionally, some traditional supervised learning methods propose domain-specific variations of color augmentation~\cite{galdran2017data,xiao2019new}.

In this paper we propose an alternative color augmentation (Figure~\ref{fig:das_visualization}, right) and we assess its impact on self-supervised learning.
We draw on the existing color imaging literature on designing features invariant to illuminant changes commonly encountered in the real world~\citep{finlayson2001solving}. Our augmentation, called \emph{Planckian Jitter}, applies physically-realistic illuminant variations. We consider the illuminants described by Planck's Law for black-body radiation, that are known to be similar to illuminants encountered in real-life~\citep{tominaga1999color}. The aim of our color augmentation is to allow the representation to contain valuable information about the surface reflectance of objects -- a feature that is expected to be important for a wide range of downstream tasks. Combining such a representation with the already high-quality shape and texture representation learned with standard data augmentation leads to a more complete visual descriptor that also describes color.

Our experiments show that self-supervised representations learned with Planckian Jitter are robust to illuminant changes. In addition, depending on the importance of color in the dataset, the proposed Planckian jitter outperforms the default color jitter. Moreover, for all evaluated datasets the combination of features of our new data augmentation with standard color jitter leads to significant performance gains of over 5\% on several downstream classification tasks. Finally, we show that Planckian Jitter can be applied to several state-of-the-art self-supervised learning methods.

\section{Background and related work}

\minisection{Self-supervised learning and contrastive learning.} Recent improvements in self-supervision learn semantically rich feature representations without the need for labelled data. In SimCLR~\citep{chen2020SimCLR} similar samples are created by augmenting an input image, while dissimilar are chosen randomly~\citep{chen2020SimCLR}. To improve efficiency, MoCo~\citep{he2020MOCO} and its enhanced version~\citep{chen2020MOCOv2} use a memory bank for learned embeddings which makes sampling efficient. This memory is kept in sync with the rest of the network during training via a momentum encoder. Several methods do not rely on explicit contrastive pairs. BYOL uses an asymmetric network incorporating an additional MLP predictor between the outputs of the two branches~\citep{NEURIPS2020BYOL}. One of the branches is kept ``offline'' and is updated by a momentum encoder. SimSiam defines a simplified solution without a momentum encoder~\citep{chen2021SimSiam}. It obtains similar high-quality results and does not require a large minibatch size, in contrast to other methods. 

\begin{figure}
    \centering
    \includegraphics[width=\linewidth]{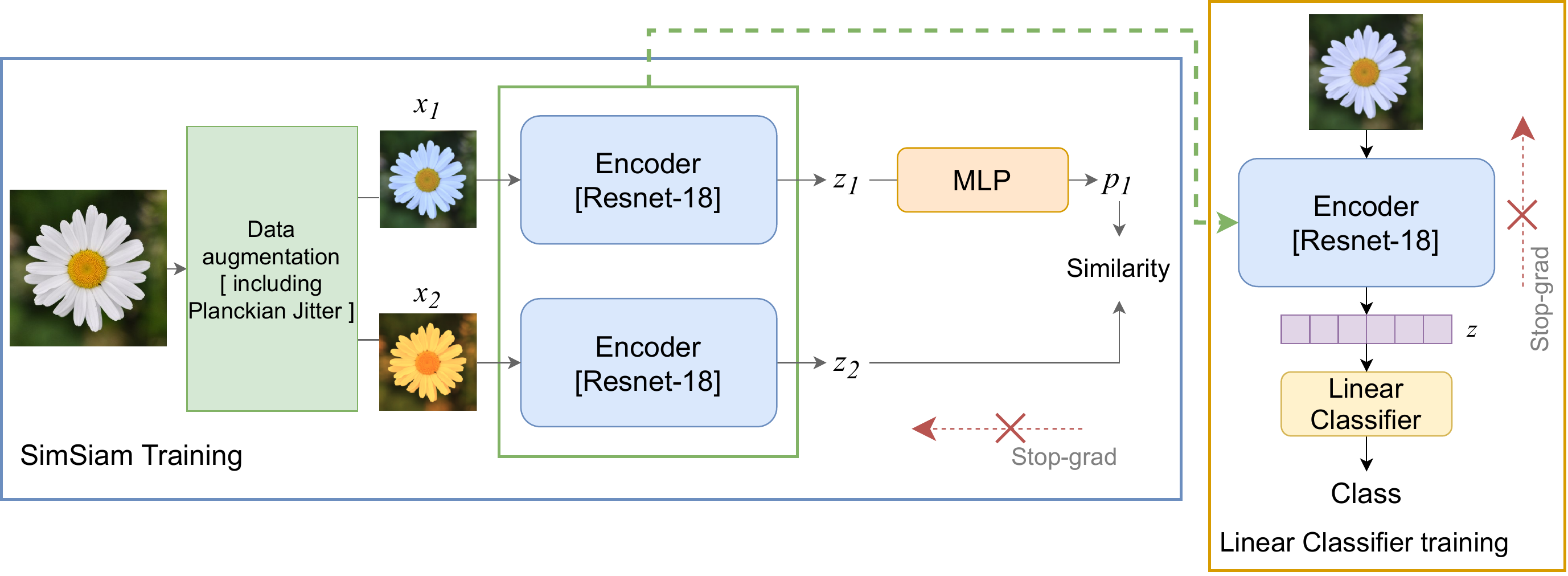}
    \caption{SimSiam training procedure exploiting Planckian-based data augmentation (left), and fine-tuning the linear classifier using the trained encoder (right).}
    \label{fig:arch}
\end{figure}

We use the SimSiam method to verify our proposed color augmentation (we also apply it to SimCLR~\citep{chen2020SimCLR} and Barlow Twins~\citep{zbontar2021barlow} in the experiments). The main component of the network is a CNN-based asymmetric siamese image encoder. One branch has an additional MLP predictor whose output aims to be as close as possible to the other (Figure~\ref{fig:arch}). The second branch is not updated during backpropagation. A negative cosine loss function is used: 
\newcommand{\dist}{\mathcal{D}}
\newcommand{\p}{{p}}  
\newcommand{\z}{{z}}  
\newcommand{\lnorm}[1]{\frac{#1}{\left\lVert{#1}\right\rVert _2}}
\newcommand{\lnormv}[1]{{#1}/{\left\lVert{#1}\right\rVert _2}}
\begin{eqnarray}
    \mathcal{L} &=& \frac{1}{2} \left[ \dist(\p_1, \text{stopgrad}(\z_2)) + \dist(\p_2, \text{stopgrad}(\z_1)) \right]\\
\label{eq:loss_simsiam}
\dist(\p_A, \z_B) &=& - \lnorm{\p_A}{\cdot}\lnorm{\z_B},
\label{eq:dist_cosine}
\end{eqnarray}
where $z_1$, $z_2$ are representations for two different augmented versions, $x_1$ and $x_2$, of the same image $x$.

The MLP is applied in alternation to either $z_1$ or $z_2$, producing respectively $p_1$ or $p_2$. Note that Figure~\ref{fig:arch} only shows an instance for $x_1$ and does not show $p_2$.
The $\text{stopgrad}(\cdot)$ operation blocks the gradient during the backpropagation. In SimSiam no contrastive term is used and only similarity is enforced during learning.

\minisection{Data augmentation.}
Data augmentation plays a central role in the self-supervised learning process. Authors \cite{chen2020SimCLR} and \cite{zbontar2021barlow} discuss the importance of the different data augmentations.
A set of well-defined transformations was proposed for SimCLR~\citep{chen2020SimCLR}. This set is commonly accepted and used in several later works. The augmentations include: rotation, cutout, flip, color jitter, blur and Grayscale. These operations are randomly applied to an image to generate the different views $x_1$, $x_2$ from which are extracted the features $z_1$ and $z_2$ used in the self-supervision loss in Eq.~\ref{eq:loss_simsiam}. Applied to the same image, contrastive-like self-supervision learns representations invariant to such distortions.

This multiple view creation is task-related~\citep{Tian2020whatmakes}, however color jittering operating on hue, saturation, brightness and contrast, is one of the most important ones in terms of overall usefulness of the learned representation for downstream tasks~\citep{chen2020SimCLR, zbontar2021barlow}. Color jitter induces a certain level of color invariance (invariance to hue, saturation, brightnesss and contrast) which are consequently transferred to the downstream task. As a consequence, we expect these learned features to underperform on downstream tasks for which color is crucial.
\cite{xiao2020should} were the first to point out that the imposed invariances might not be beneficial for downstream tasks. As a solution, they propose to learn different embedding spaces in parallel that capture each of the invariances. Differently than them, we focus on the color distortion and propose a physics-based color augmentation that allows learning invariance to physically realistic color variations.

Color imaging has a long tradition in research on color features invariant to scene-accidental events such as shading, shadows, and illuminant changes~\citep{geusebroek2001color, finlayson2001solving}. Invariant features were found to be extremely beneficial for object recognition. The invariance to hue and saturation changes induced by color jitter, however, is detrimental to object recognition for classes in which color characteristics are fundamentally discriminative. Therefore, in this work we revisit early theory on illuminant invariance~\citep{finlayson2001solving} to design an improved color augmentation that induces invariances common in the real world and that, when used during self-supervised learning, does not damage the color quality of the learned features. 

\section{Methodology}

The image transformations introduced by default color jitter creates variability in training data that indiscriminately explores all hues at various levels of saturation. The resulting invariance is useful for downstream tasks where chromatic variations are indeed irrelevant (e.g. car color in vehicle recognition), but is detrimental to downstream tasks where color information is critical (e.g. natural classes like birds and vegetables).
The main motivation for applying strong color augmentations is that this it leads to very strong shape and texture representations. Indiscriminately augmenting color information in the image requires that the representation solve the matching problem using shape~\citep{chen2020SimCLR}\footnote{This is pointed out in the discussion of Figure 5 in~\cite{chen2020SimCLR}}. 

As an alternative to color jitter, we propose a physics-based color augmentation that mimics color variations due to illuminant changes commonly encountered in the real world. The aim is to reach a representation that does not have the color crippling effects of color jitter, and that better describes classes for which surface reflectance is a determining feature. When combined with default color jitter, this representation should also provide a high-quality shape/texture and color representation.

\subsection{Planckian Jitter}
\label{sec:PlanckianJitter}
We call our color data augmentation procedure \emph{Planckian Jitter} because it exploits the physical description of a black-body radiator to re-illuminate training images within a realistic illuminant distribution~\citep{finlayson2001solving,tominaga1999color}. The resulting augmentations are more realistic than those of the default color jitter (see Fig.~\ref{fig:das_visualization}). The resulting learned, self-supervised feature representation is thus expected to be robust to illumination changes commonly observed in real-world images, while simultaneously maintaining the ability to discriminate the image content based on color information.

Given an input RGB training image $I$, our Planckian Jitter procedure applies a chromatic adaptation transform that simulates realistic variations in the illumination conditions. The data augmentation procedure is as follows:
\begin{enumerate}
    \item we sample a new illuminant spectrum $\sigma_T(\lambda)$ from the distribution of a black-body radiator;
    \item we transform the sampled spectrum $\sigma_T(\lambda)$ into its sRGB representation $\rho_T \in \mathbb{R}^3$;
    \item we create a jittered image $I'$ by reilluminating $I$ with the sampled illuminant $\rho_T$; 
    \item we introduce brightness and contrast variation, producing a Planckian-jittered image $I''$.
\end{enumerate}

A radiating black body at temperature $T$ can be synthesized using Planck's Law~\citep{andrews2010introduction}:
\begin{equation}
    \sigma_T(\lambda) = 
    \frac{2 \pi h c^2}{\lambda^5 (e^{\frac{h c}{k T \lambda}} - 1)} \text{ W} / \text{m}^3,
\end{equation}
where $c = 2.99792458 \times 10^8$ m/s is the speed of light, $h = 6.626176 \times 10^{-34}$ Js is Planck's constant, and $k = 1.380662 \times 10^{-23}$ J/K is Boltzmann's constant.
We sampled $T$ in the interval between $3000K$ and $15000K$ which is known to result in a set of illuminants that can be encountered in real life~\citep{tominaga1999color}.
Then, we discretized wavelength $\lambda$ in 10nm steps ($\Delta\lambda$) in the interval between 400nm and 700nm.
The resulting spectra are visualized in Figure~\ref{fig:planckian_distribution} (left) in Appendix~\ref{app:planckian_distribution}.

The conversion from spectrum into sRGB is obtained according to~\cite{wyszecki1982color}:
\begin{enumerate}
    \item we first map the spectrum into the corresponding XYZ stimuli, using the 1931 CIE standard observer color matching functions $c^{\{X,Y,Z\}}(\lambda)$,
    in order to bring the illuminant into a standard color space that represents a person with average eyesight;

    \item we normalize this tristimulus by its $Y$ component, convert it into the CIE 1976 L*a*b color space, and fix its L component to 50 in a 0-to-100 scale, allowing us to constrain the intensity of the represented illuminant in a controlled manner as a separate task; and
    
    \item we then convert the resulting values to sRGB, applying a gamma correction and obtaining $\rho_T = \{R,G,B\}$; the resulting distribution of illuminants is visualized with the Angle-Retaining Chromaticity diagram~\citep{buzzelli2020arc} in Figure~\ref{fig:planckian_distribution} (right) in Appendix~\ref{app:planckian_distribution}.
\end{enumerate}

All color space conversions assume a D65 reference white, which means that a neutral surface illuminated by average daylight conditions would appear achromatic. Once the new illuminant has been converted in sRGB, it is applied to the input image $I$ by resorting to a Von-Kries-like transform~\citep{von1902theoretische} given by the following channel-wise scalar multiplication:
\begin{equation}
    I'^{\{R,G,B\}} = I^{\{R,G,B\}} \cdot \{R,G,B\} / \{1, 1, 1\},\end{equation}
where we assume the original scene illuminant to be white (1,1,1).
Finally, brightness and contrast perturbations are introduced to simulate variations in the intensity of the scene illumination:
\begin{equation}
    I'' =  c_B \cdot c_C\cdot I'+ \left(1-c_C\right)\cdot\mu \left(c_B \cdot I' \right),
\end{equation}
where $c_B=0.8$ and $c_C=0.8$ represent, respectively, brightness and contrast coefficients, and $\mu$ is a spatial average function. 

\subsection{Complimentarity of shape, texture and color representations}
\label{sec:feature_concat}

The self-supervised learning paradigm involves a pretraining phase that relies on data augmentation to produce a set of features with certain invariance properties. These features are then used as the representation for a second phase, where we learn a given supervised downstream task. The default color jitter augmentation generates features that are strongly invariant to color information, resulting in high-quality representations of shape and texture, but that is an inferior descriptor of surface reflectances (i.e. the color of objects). Our augmentation based on Planckian Jitter (see Figure~\ref{fig:das_visualization}) is based on transformations mimicking the physical color variations in the real world due to illuminant changes. As a result, the learned representation yields a high-quality color description of scene objects (this is also verified in Appendix~\ref{app:color_importance}). However, it  likely leads to a drop in the quality of the shape and texture representation (since color can be used to solve cases where previously shape/texture were required). To exploit the complimentarity of the two representations, we propose to learn both -- one with color jitter and one with Planckian Jitter -- and to then concatenate the results in a single representation vector (of 1024 dimensions, i.e. twice the original size of 512). We call this \textit{Latent space combination (LSC)}.


\section{Experimental results}
In this section, we analyze the color sensitivity of the learned backbone networks, verify the superiority of the proposed color data augmentation method compared to the default color jitter on color datasets, and evaluate the impact on downstream classification tasks. We report additional results on computational time of the proposed Planckian augmentation in Appendix~\ref{app:time}.

\subsection{Training and evaluation setup}
\label{sec:train_setup}
We perform unsupervised training on two datasets: CIFAR-100~\citep{krizhevsky2009learning} ($32\times 32$) and ImageNet ($224\times 224$).
\footnote{We conduct the investigative part of our research in an agile manner on low-resolution images, then transfer the most significant configurations to a higher-resolution, to ether confirm or refute the initial hypotheses. 
}
We slightly modify the ResNet18 architecture to accommodate $32 \times 32$ images: the kernel size of the first convolutional was reduced from $7\times7$ to $3\times3$ and the first max pooling layer was removed. SimSiam training was performed using Stochastic Gradient Descent with a starting learning rate of $0.03$, a cosine annealing learning rate scheduler, and mini-batch size of 512 (as in original SimSiam work by~\cite{chen2021SimSiam}). For the training on the 1000-class ImageNet training set, we follow the same procedure as~\cite{chen2021SimSiam} with ResNet50.

The linear classifier training at resolution $32\times32$ was performed on \textsc{Cifar-100} and \textsc{Flowers-102}~\citep{nilsback2008automated}. \textsc{Cifar-100} is used as a baseline for the classification task. The linear classifier training for \textsc{Cifar-100} is done with Stochastic Gradient Descent for 500 epochs with a starting learning rate $0.1$, a cosine annealing learning rate scheduler, and mini-batch size of 512.
The \textsc{Flowers-102} dataset with 102 classes was selected to assess the quality of the features extracted in scenarios where color information plays an important role. Images from \textsc{Flowers-102} are resized to $32\times32$ pixels to match the input dimensions of the pretrained model. Here we used the Adam optimizer with initial learning rate of $0.03$.

For training linear classifiers at resolution $224\times224$ for downstream tasks we follow the evaluation protocol of~\cite{chen2021SimSiam}.
We use six different datasets: \textsc{ImageNet}, \textsc{Flowers-102}, the fine-grained \textsc{VegFru}~\citep{Hou2017VegFru}, \textsc{CUB-200}~\citep{WelinderEtal2010}, \textsc{T1K+}~\citep{cusano2021t1k+}, and \textsc{USED}~\citep{ahmad2016used}, all resized to $224\times224$ pixels. More details about the datasets are provided in Appendix~\ref{app:datasets}.
In the case of \textsc{CUB-200}, each image was cropped using the bounding boxes given in the dataset annotations. For \textsc{T1K+}, we used the 266 class labeling to train and test the linear classifier.

To assess the impact of color data augmentations we define six different configurations:
\begin{itemize}
    \item \textit{Default Color Jitter (CJ):}
    the default configuration, as used in SimSiam and SimCLR, uses both Random Color Jitter and Random Grayscale operations. 
    \item \textit{Default Color Jitter w/o  Grayscale (CJ-)}: same as \textit{Default}, without Random Grayscale.
    \item \textit{Planckian Jitter (PJ)}: uses the complete proposed Planckian Jitter operating on chromaticy, brightness, and contrast aspects of the images. No Random Grayscale is applied.
    \item \textit{LSC Default Color Jitter + Planckian Jitter ([CJ,PJ]}: This latent space combination (simple concatenation of representations) combines the default color jitter with our Planckian jitter. It allows evaluation of the complimentary nature of the representations.
    \item \textit{LSC Default Color Jitter + Default Color Jitter w/o Grayscale ([CJ,CJ-])}: We combine the default color jitter with a version without the Grayscale augmentation, since this representation is also expected to result in a better color representation. 
    \item \textit{LSC of two Default Color Jitter Models ([CJ,CJ])}: We also show results of simply concatenating two independently trained models (trained from different seeds) with default color jitter (an ensemble of two models). 
\end{itemize}
In all experiments, these are combined with the other default augmentations (crop, flip, and blur).

\subsection{Color sensitivity analysis}
\label{sec:color_sensitivity}

\begin{figure}
\setlength{\tabcolsep}{1pt}
\begin{tabular}{cc}
\includegraphics[width=.65\linewidth]{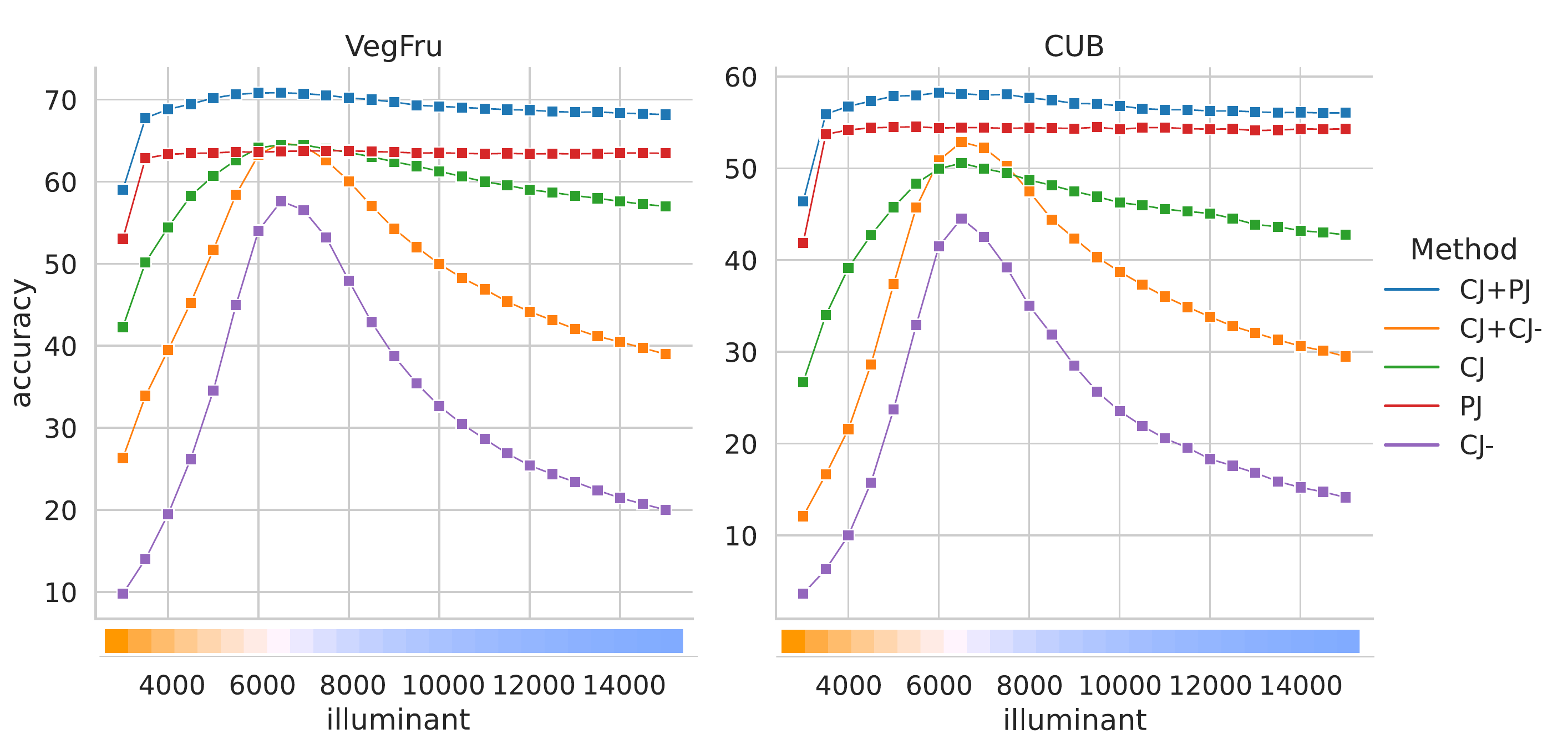} &
\includegraphics[width=.35\linewidth]{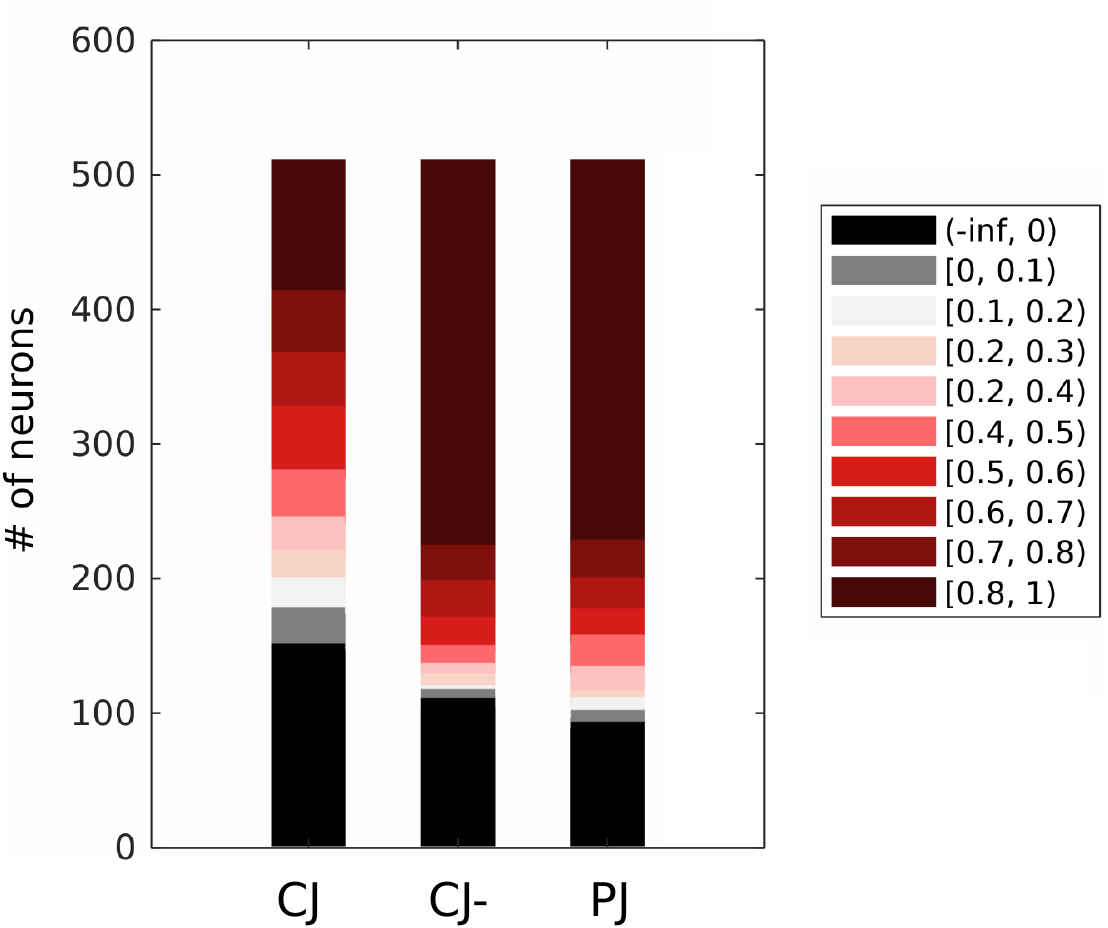} \\
(a) & (b)
\end{tabular}
\caption{Color sensitivity analysis. (a) Robustness to illuminant change: we report the accuracies by differently-trained backbones as a function of illuminant. (b) The color sensitivity indexes computed for the different configurations used for training the backbone.}
\label{fig:color_analysis1}

\end{figure}

We perform a robustness analysis on the VegFru and \textsc{CUB-200} datasets with realistic illuminant variations, and analyzed sensitivity to color information.
This experiment is driven by two motivations: to verify that we obtain invariance to the transformation applied during training, and to characterize the degradation of different non-Planckian training modalities.
We assume as reference point the D65 illuminant, which for the purpose of this test is considered the default illuminant in every image. Given the different backbones pretrained on \textsc{ImageNet}, we then train a linear classifier on this dataset (assumed to be under white illumination). For testing we  create different versions of the dataset, each illuminated by illuminants of differing color temperature. This allows us to evaluate the robustness of the learned representations with respect to these illuminant changes.

Results are given in Figure~\ref{fig:color_analysis1}(a) (more results are provided in Figure~\ref{fig:pj_analysis} from Appendix~\ref{app:color_sens}). \emph{Planckian Jitter} obtains a remarkably stable performance between 4000K and 14000K, while \textit{Default Color Jitter} is more sensitive to the illumination color and the classification accuracy decreases when the scene illuminant moves away from white. We also see that the combination of default and Planckian Jitter obtains the best results for all illuminants and manages to maintain a high-level of invariance with respect to the illuminant color. Among the non-Planckian curves, default color jitter (CJ) is the most invariant, followed by CJ+CJ- (although showing better performance at D65), and finally CJ-.

In order to understand the impact of the color information on each neuron in trained models, we conducted an analysis using the color selectivity index described by~\cite{rafegas2018color}.
This index measures neuron activation when color is present or absent in input images. We computed the index for the last layer of different backbones, and high values indicate color-sensitive neurons. See Appendix~\ref{app:color} for more details on color selectivity. The results are shown in Figure~\ref{fig:color_analysis1}(b) and indicate the number of color-sensitive neurons for each of the considered models.
It is clear that the default color jitter has far fewer neurons dedicated to color description. This result confirms the hypothesis that models trained in this way are color invariant, a property that negatively affects the model in scenarios where color information has an important role as seen in our experiments. We have also analyzed the results for the default color jitter without Grayscale augmentation (CJ-). These results show that removing the Grayscale augmentation improves color sensitivity significantly. We therefore also consider this augmentation in future experiments.

\subsection{Ablation study}

\begin{table*}[t]
\centering
\caption{Accuracy results with ablation on color augmentations. Self-supervised training is performed on \textsc{Cifar-100} and the learned features are evaluated at ($32\times32$) on \textsc{Cifar-100} and \textsc{Flowers-102}. Augmentation techniques include variations in hue and saturation (H\&S), brightness and contrast (B\&C), Planckian-based chromaticity (P), and random Grayscale conversions (G). Accuracy refers to the linear classifiers trained with features extracted from the different backbones.}
\begin{center}
\begin{tabular}{lccccrr}
\textbf{AUGMENTATION} & \textbf{H\&S} & \textbf{B\&C} & \textbf{G} & \textbf{P} & \textbf{\textsc{Cifar-100}} & \textbf{\textsc{Flowers-102}} \\
\hline\\[-1.8ex]
None &  &  &  &  & 41.93\% & 36.47\% \\
Default Color Jitter & \checkmark & \checkmark & \checkmark &  & 59.93\% & 30.00\% \\
 & \checkmark & \checkmark &  &  & 41.96\% & 36.96\% \\
 & \checkmark &  &  &  & 32.46\% & 39.11\% \\
 &  &  &  & \checkmark & 36.10\% & 39.51\% \\
 &  & \checkmark &  &  & 31.78\% & 41.96\% \\
Planckian Jitter &  & \checkmark &  & \checkmark & 47.31\% & 42.75\% \\
\hline
\end{tabular}
\end{center}
\label{tab:augmentations_res}

\end{table*}

\begin{table*}[t]
\centering
\caption{Accuracy results for self-supervised training on \textsc{Cifar-100} and evaluated at $32\times32$ on \textsc{Cifar-100} and \textsc{Flowers-102}. The reported accuracy refers to the results of the linear classifiers trained with features extracted from the different trained backbones.}
\begin{center}
\begin{tabular}{lrr}
\textbf{AUGMENTATION} & \textbf{\textsc{Cifar-100}} & \textbf{\textsc{Flowers-102}} \\
\hline\\[-1.8ex]
Default Color Jitter (CJ) & 59.93\% & 30.00\% \\
Default Color Jitter w/o Grayscale (CJ-) & 41.96\% & 36.96\% \\
Planckian Jitter (PJ) & 47.31\% & 42.75\%\vspace{1.5mm} \\
LSC: [CJ, CJ-] & 62.27\% & 47.65\% \\
LSC: [CJ, PJ] & 63.54\% & 51.66\% \\
\hline
\end{tabular}
\end{center}
\label{tab:aug_cifar_res}
\vspace{-3mm}
\end{table*}

Six different models were trained and evaluated with a linear classification for image classification. For resolution $32\times32$ the model is evaluated on \textsc{Cifar-100} and \textsc{Flowers-102}. The results in terms of accuracy are reported in Table~\ref{tab:augmentations_res} and Table~\ref{tab:aug_cifar_res}.
We identify two different trends when interpreting these results. On \textsc{Cifar-100},  removing color augmentations  makes the model less powerful, due to the loss of color invariance in the features extracted by the encoder. This behaviour is consistent with what was reported by~\cite{chen2020SimCLR}.
We see in Table~\ref{tab:augmentations_res} that if color augmentations (i.e. brightness/contrast and Random Grayscale) are removed completely (the \textit{None} configuration), the accuracy drops by $18\%$.
On \textsc{Flowers-102} the behavior is the opposite however: removing color augmentations helps the model to better classify images, obtaining an improvement of $12.75\%$ of accuracy with respect to the default color jitter.
This behavior confirms that color invariance negatively impacts downstream tasks where color information plays an important role.

Taking a closer look at the various augmentation on \textsc{Flowers-102}, we see that introducing more realistic color augmentations positively impacts contrastive training and produces models that achieve even better results with respect to the configuration without any kind of image color manipulation.
Removing all color augmentations (None) improves results already by over 6\%. Then, by simply reducing the jittering operation to influence brightness and contrast, leaving hue and saturation unchanged, yields another boost in accuracy of $5.49\%$ (to 41.96). When we start modifying chromaticity using a more realistic transformation (i.e \textit{Planckian Jitter}), the final result is a boost of $6.28\%$ in accuracy with respect to the \textit{None} configuration. Also, on \textsc{Cifar-100} we see an improvement of $5.38\%$ from Planckian Jitter with respect no color augmentation. Despite this improvement, in this scenario the contrastive training with the realistic augmentation does not yield better results with respect to the \textit{Default} configuration because color only plays a minor role on this dataset.

Given the results obtained using the data augmentations reported in Table~\ref{tab:augmentations_res}, and given the considerations made in Section~\ref{sec:feature_concat}, we evaluate the complementarity of the learned representation by combining latent spaces from different backbones. Results for two different latent space combinations are given in Table~\ref{tab:frameworks_and_mix}. On both datasets the \textit{Latent space combination} of Default and Planckian Jitter achieves the best results. On the original \textsc{Cifar-100} task, this combination achieves a total accuracy of $63.54\%$, a $3.61\%$ improvement over \textit{Default} and $16.23\%$ more compared to \textit{Planckian Jitter} alone. Comparing to the LSC using the Default ColorJitter w/o Grayscale, the version with Planckian Jitter achieves a small improvement of $1.27\%$ in classification accuracy.

On the downstream \textsc{Flowers-102} task, \textit{Latent space combination} reaches an accuracy value of $51.66\%$: an improvement of $21.66\%$ and $8.91\%$ in accuracy respectively compared to the two original configurations.
Compared to the LSC using Default ColorJitter w/o Grayscale, the combination with Planckian Jitter achieves a higher result, and a bigger gap in terms of accuracy with respect to the \textsc{Cifar-100} scenario. Here the use of Planckian Jitter brings an improvement of $4.01\%$, confirming the impact of using realistic augmentation on classification tasks for which color is important.

\subsection{Evaluation on downstream tasks}
\label{sec:downstream_tasks}
Given the ablation study results, we performed the analysis of the proposed configurations on other downstream tasks using the backbone trained on higher resolution images ($224\times224$ pixels).
We report in Table~\ref{tab:imagenet} the results for:  \textit{Default Color Jitter}, \textit{Planckian Jitter}, and latent space combinations.

Looking at the results, we see that the \textit{Planckian Jitter} augmentation outperforms default color jitter on three datasets (\textsc{CUB-200}, \textsc{T1K+}, and \textsc{USED}). Comparing the results on \textsc{Flowers-102} with those reported above at $(32\times 32$) pixels, we see that default color jitter actually obtains good results. We hypothesize that for high-resolution images the shape/texture information is very discriminative, and the additional color information yields little gain (for further analysis, see also~\ref{app:resolution_importance}).
Table~\ref{tab:imagenet} also contains results for latent space combination, which confirm that the two learned representations are complementary. Their combination yields gains of up to 9\% on T1K+. As a sanity check we also include the latent space combination of two networks separately trained with Color Jitter. This provides a small gain on some datasets, but yields significantly inferior results than LSC.

\begin{table}[]
\centering
\caption{Evaluation on downstream tasks. Self-supervised training was performed on \textsc{ImageNet} at ($224\times224$) and testing performed on the downstream datasets resized to ($224\times224$).}
\label{tab:imagenet}
\begin{tabular}{lrrrrrr}
\bf AUGMENTATION  & \multicolumn{1}{l}{\bf \textsc{CUB-200}} & \multicolumn{1}{l}{\bf \textsc{VegFru}} & \multicolumn{1}{l}{\bf \textsc{T1K+}} & \multicolumn{1}{l}{\bf \textsc{USED}} & \multicolumn{1}{l}{\bf \textsc{Flowers-102}}\\\hline\\[-1.8ex]
Default Color Jitter (CJ)           & 54.52\% & 67.63\% & 71.44\% & 59.90\% & 93.16\% \\
Planckian Jitter (PJ)               & 56.28\% & 65.84\% & 77.42\% & 60.03\% & 90.29\% \\
LSC [CJ,PJ]                         & \bf{60.70}\% & \bf{74.73}\% & \bf{80.49}\% & \bf{64.07\%} & \bf{93.99}\% \\
LSC [CJ,CJ]   & 56.16\% & 70.59\% & 73.47\% & 61.07\% & 93.13\% \\
LSC [CJ,CJ-]    & 53.14\% & 70.54\% & 78.32\% & 63.87\% & 93.47\% \\
\hline
\end{tabular}
\vspace{-3mm}
\end{table}

\begin{table}[]
\centering
\caption{Effect of Plackian Jitter on different contrastive learning models. Self-supervised training was performed on \textsc{CIFAR-100} and the learned features are evaluated at ($32\times32$) on \textsc{CIFAR-100} and \textsc{Flowers-102}. We report the best configurations obtained on SimSiam model and retrained SimCLR and Barlow Twins with those selected configurations.}
\label{tab:frameworks_and_mix}
\begin{tabular}{llrr}
\bf FRAMEWORK                       & \bf AUGMENTATION                                       & \multicolumn{1}{l}{\bf \textsc{Cifar-100}} & \multicolumn{1}{l}{\bf \textsc{Flowers-102}} \\\hline\\[-1.8ex]
\multirow{3}{*}{SimSiam}        & Default Color Jitter (CJ)     & 59.93\%                      & 30.00\%                         \\
                                & Planckian Jitter (PJ)                                   & 47.31\%                      & 42.75\%                         \\
                                & LSC [CJ,PJ]                              & 63.54\%                      & 51.66\%                         \\\hline\\[-1.8ex]
\multirow{3}{*}{SimCLR}         & Default Color Jitter (CJ)     & 56.99\%                      & 35.29\%                         \\
                                & Planckian Jitter (PJ)                                   & 47.75\%                      & 45.00\%                         \\
                                & LSC [CJ,PJ]                              & 61.07\%                      & 55.78\%                         \\\hline\\[-1.8ex]
\multirow{3}{*}{Barlow Twins}   & Default Color Jitter (CJ)     & 56.60\%                      & 40.78\%                         \\
                                & Planckian Jitter (PJ)                                   & 52.71\%                      & 54.50\%                         \\
                                & LSC [CJ,PJ]                             & 62.85\%                      & 62.55\%                         \\\hline\\[-1.8ex]
\multirow{3}{*}{VicReg}   & Default Color Jitter (CJ)     & 65.23\%                      & 49.50\%                         \\
                                & Planckian Jitter (PJ)                             & 59.19\%                      & 50.90\%                         \\
                                & LSC [CJ,PJ]                             & 68.95\%                      & 60.80\%                         \\\hline
\end{tabular}
\vspace{-2mm}
\end{table}

\subsection{Generality and Limitations of Planckian Jitter}
To show that our approach is generally applicable to self-supervised methods exploiting color augmentations, we report in Table~\ref{tab:frameworks_and_mix} experiments using SimCLR, Barlow Twins, and the more recent VicReg~\cite{bardes2021vicreg}.
Independently of the model, \textit{Latent Space Combination} consistently achieves the best results on both datasets.

A drawback of Planckian Jitter is the quality reduction of shape and texture representations, because the extreme color transformation of the standard Color Jitter force the network to solve the contrastive learning problem mainly using shape/texture information. As we have shown, this problem can be addressed by exploiting their complimentary nature. Secondly, our current latent space combination requires the training of two separate backbones, which will also learn partially-overlapping features. A training scenario with both augmentations simultaneously in a single network while reserving part of the latent space for each augmentation could be pursued to address this limitation. 
Finally, object-specific augmentations that also take into account shadows, the type of reflectance, secondary light sources, inter-reflections, shadows, etc, could lead to further improvements.

\section{Conclusion}
Existing research on self-supervised learning mainly focuses on tasks where color is not a decisive feature, and consequently exploits data augmentation procedures that negatively affect color-sensitive tasks.
We propose an alternative color data augmentation, called Planckian Jitter, that is based on the physical properties of light. Our experiments demonstrate its positive effects on a wide variety of tasks where the intrinsic color of the objects (related to their reflectance) is crucial for discrimination, while the illumination source is not. 
We also proposed exploiting both color and shape information by concatenating features learned with different modalities of self-supervision, leading to significant overall improvements in learned representations. Planckian Jitter can be easily incorporated into any self-supervised learning pipeline based on data augmentations, as shown by our results demonstrating improved performance for three self-supervised learning models.

\subsubsection*{Reproducibility Statement}

The code of the Planckian Jitter data augmentation procedure, written in MATLAB and PyTorch 1.7.0, will be made available upon acceptance.

The training runs have been performed using Pytorch  in combination with Pytorch Lightning Bolt framework, which provides an implementation of SimSiam methodology for backbone contrastive training. The model has been trained using CUDA deterministic, and random seed set to $1234$.

All datasets used for the training and fine-tuning are publicly available. Only the CUB200 dataset has been pre-processed, by cropping each image using the given bounding box values, available alongside each image in the annotations files.

\subsubsection*{Acknowledgments}
We acknowledge the support from the Spanish Government funding for projects PID2019-104174GB-I00, TED2021-132513B-I00, the support from the European Commission under the Horizon 2020 Programme, grant RYC2021-032765-I funded by MCIN/AEI/10.13039/501100011033 and by European Union NextGenerationEU/PRTR, and grant number 951911 (AI4Media).

\bibliographystyle{plainnat}
\bibliography{egbib}


\newpage
\appendix
\section{Appendix A}

\subsection{Planckian Jitter}
\label{app:planckian_distribution}

\begin{figure}[h]
    \centering
    \includegraphics[width=\columnwidth]{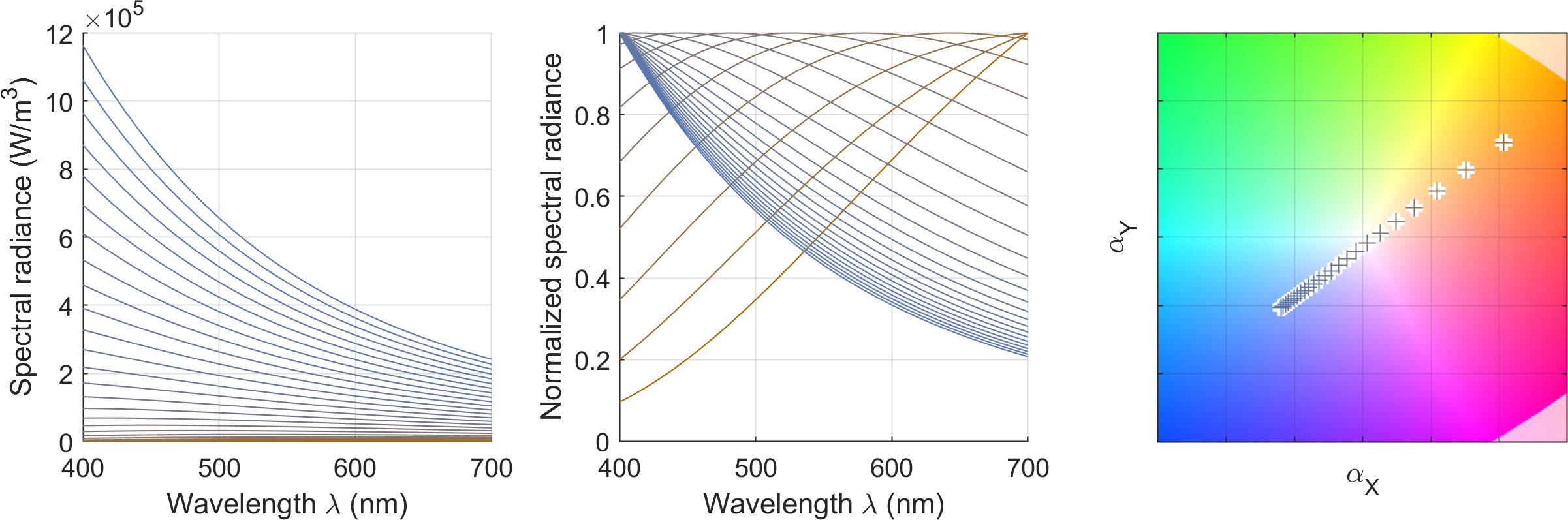}
    \caption{Spectral power distributions (left) and corresponding ARC chromaticities (right) of the sampled black body radiator, used to generate Planckian jittering.}
    \label{fig:planckian_distribution}
\end{figure}

Figure~\ref{fig:planckian_distribution} illustrates the illuminants sampled from the distribution of a black body radiator, with correlated color temperature $T$ in the interval between $3000K$ and $15000K$.
The resulting spectra are visualized on the left and in the middle, while the resulting distribution of illuminants is visualized in the Angle-Retaining Chromaticity diagram on the right.

\begin{figure}[t]
    \centering
    \includegraphics[width=.8\columnwidth]{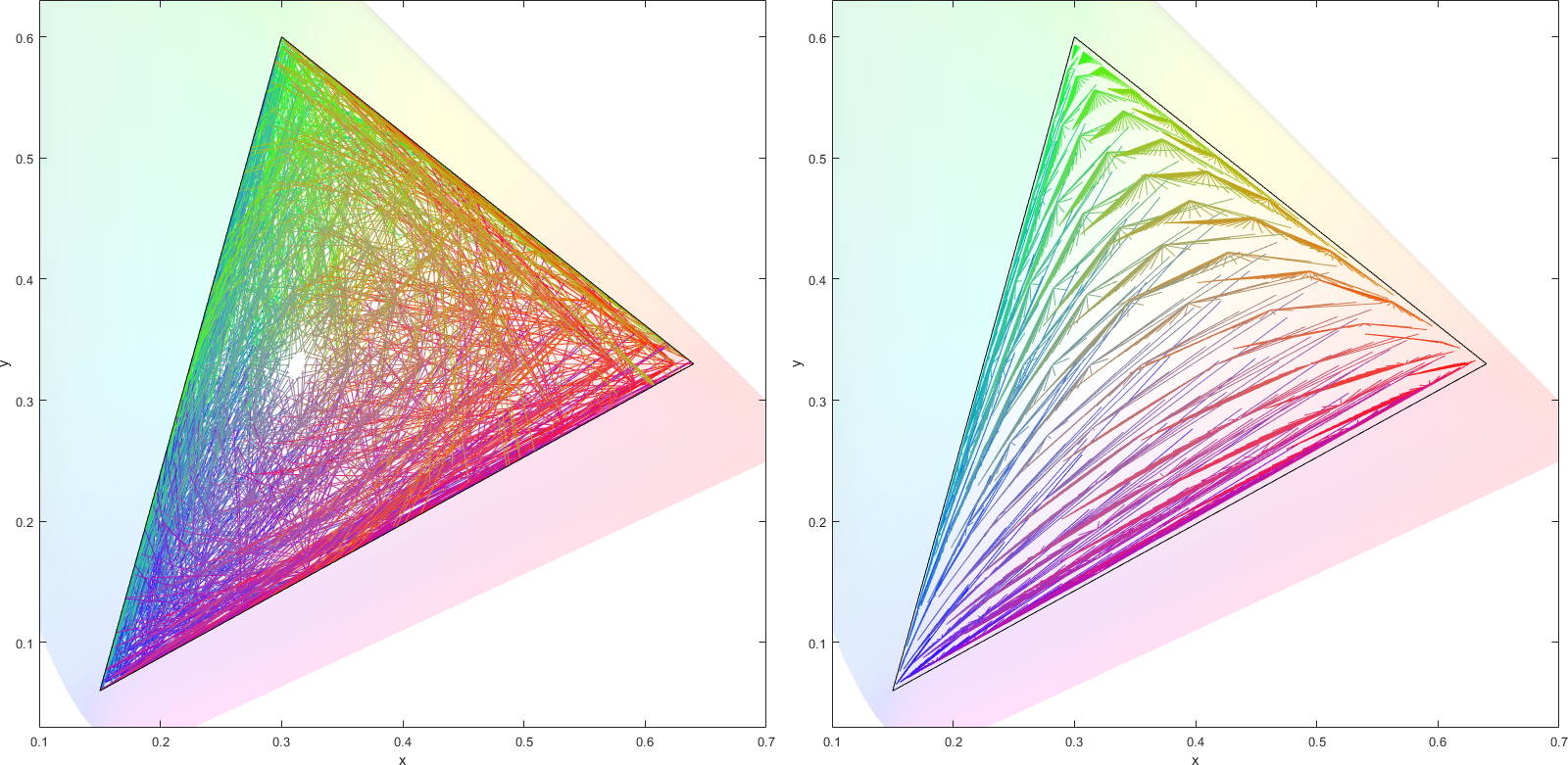}
    \caption{Default color jitter (left) and Planckian jitter (right) in xy chromaticity.}
    \label{fig:das_visualization_xy}
\end{figure}

Figure~\ref{fig:das_visualization_xy} shows a comparison between default color jitter (left) and Planckian jitter (right), replicating Figure 1 in xy chromaticity.

\subsection{Dataset details}
\label{app:datasets}
\begin{figure}[h]
    \centering
    \includegraphics[width=\linewidth]{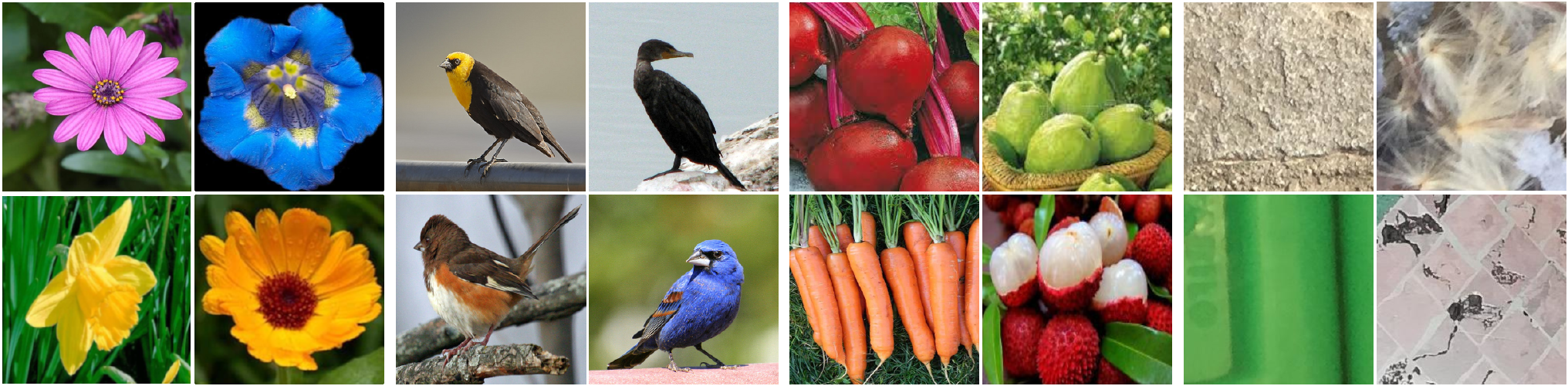}
    \caption{Example images from the datasets used as downstream classification tasks. From left to right: \textsc{Flowers-102}, \textsc{CUB-200}, \textsc{VegFru}, and \textsc{T1K+}.}
    \label{fig:datasets}
\end{figure}

In section 4.4 of the main paper we analyzed the impact of our data augmentation when using the features extracted from the backbone trained on \textsc{ImageNet} on new datasets. The datsets used in the finetuning step are:
\begin{itemize}
    \item \textsc{Flowers-102}~\citep{nilsback2008automated}: Dataset consisting of 102 flower categories commonly occurring in the United Kingdom. Each class consists of between 40 and 258 images, for a total 8,189 images. 
    \item \textsc{VegFru}~\citep{Hou2017VegFru}: Dataset consisting of more than 160,000 images of vegetables and fruits divided in 292 classes.
    \item \textsc{CUB-200}~\citep{WelinderEtal2010}: Dataset made of 6,033 images of 200 bird species. 
    \item \textsc{T1K+}~\citep{cusano2021t1k+}: Dataset of textures divided into 1129 classes and organized in 5 groups of 266 super classes. We adopted the 266 class labeling to finetune and test our models.
    \item \textsc{USED}~\citep{ahmad2016used}: Dataset consisting of 14 categories of social events from around the world. Images depict the interaction between multiple objects and the background scene. We considered 1000 images per class for training, and 500 images per class for testing.
\end{itemize}
A few example images for each of the color task datasets are given in Figure~\ref{fig:datasets}.

Additionally, in section~\ref{app:extratiny} of this appendix \textsc{Tiny-ImageNet}~\citep{le2015tiny} is used. It contains 100,000 images of 200 classes (500 for each class) at $64\times64$ pixel resolution.

\subsection{Color selectivity index}
\label{app:color}

\begin{figure}[t]
    \centering
    \includegraphics[width=\linewidth]{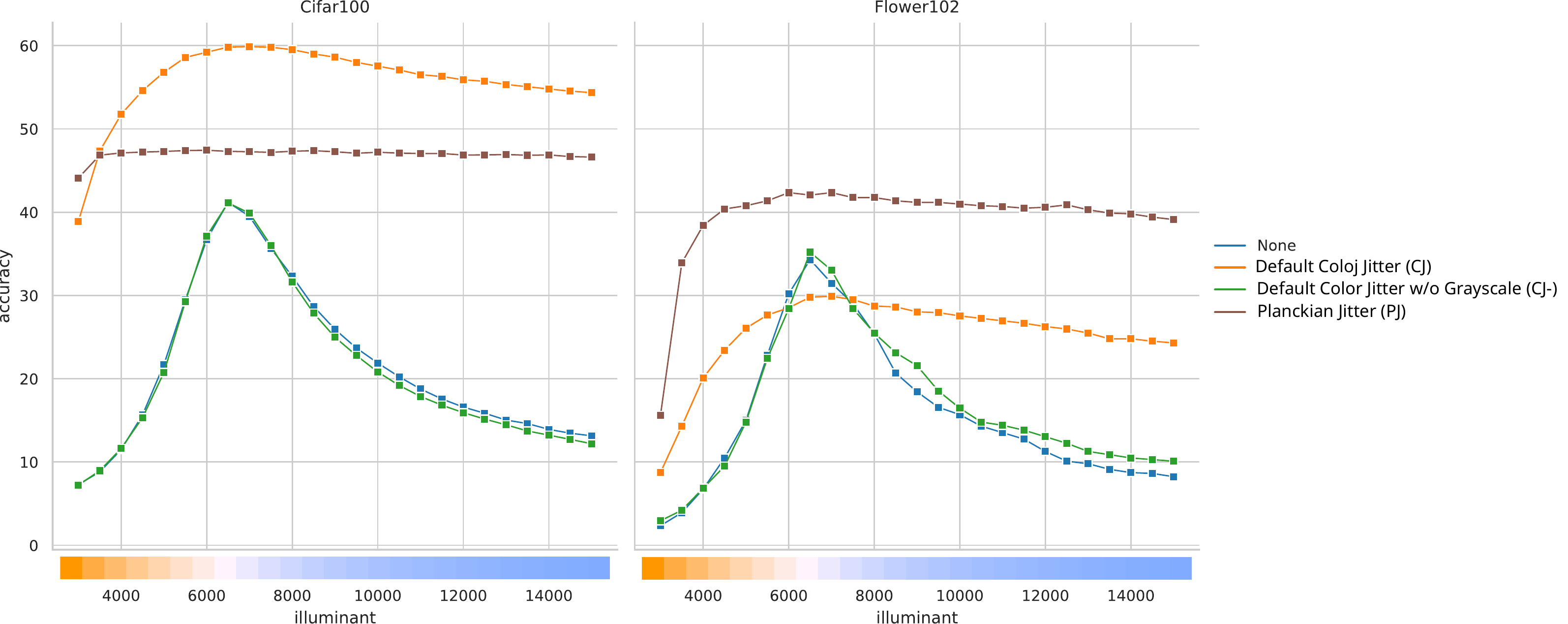}
    \caption{Illumimant robustness analysis. To assess feature invariance to realistic color changes in images, for each method we evaluate classification accuracy on 25 different, re-illumintated versions of the datsets. The images of the two datasets (\textsc{Cifar-100} on the left and \textsc{Flowers-102} on the right) have been modified with the illumimants from temperature 3000 K to 15000 K using the Planckian Jitter transform.}
    \label{fig:pj_analysis}
\end{figure}

Color selectivity is defined by~\cite{rafegas2018color} as the property of a neuron that activates strongly when a specific color appears in the input image, and does not when the color is absent. It is computed by estimating the ratio between the neuron's global activation with color input images and the global activation with corresponding grayscale images:
\begin{equation}
    \alpha(n^{L,i}) = 1-\frac{\sum\limits_{j=1}^{N} w'_{j,i,L}}{\sum\limits_{j=1}^{N} w_{j,i,L}}.
\end{equation}
Here $w_{j,i,L}$ refers to the activation of an image patch $j$ for the $i$-th neuron $n^{L,i}$ at layer $L$,
normalized for the maximum activation value across all possible image patches.
$w'_{j,i,L}$ is the equivalent formulation for a grayscale version of the images.
The set of considered image patches is restricted to the top-$N$ regions from a given dataset that maximally activate the neuron of interest.

We can distinguish between neurons that are colorblind or neurons that highly rely on color information by looking at the $\alpha$ value obtained: an $\alpha$ value more than $0.25$ means that the neuron is high color selective, while an alpha value less than $0.1$ means that the neuron is basically colorblind. These thresholds were selected based on the analysis by~\cite{rafegas2018color}.
We collected alpha values for the neurons in the last layer of the encoders trained with different data augmentation configurations in order to compare the models sensitivity to color and how it changes in relation to the training procedure adopted.

\subsection{Color sensitivity}
\label{app:color_sens}

To analyze feature robustness to different illuminants, we tested the models with different, re-illuminated versions of the \textsc{Cifar-100} and \textsc{Flowers-102} datasets.
We applied \textit{Planckian Jitter} on the two datasets, generating 25 different versions of each, one for each illuminant sampled.
Using these different versions of the datasets we then test the models for each illuminant and collect the classification accuracies. The results on both \textsc{Cifar-100} and \textsc{Flowers-102} are given in Figure~\ref{fig:pj_analysis}.

\subsection{Execution time comparison}
\label{app:time}
\begin{figure}[t]
    \centering
    \includegraphics[width=.6\columnwidth]{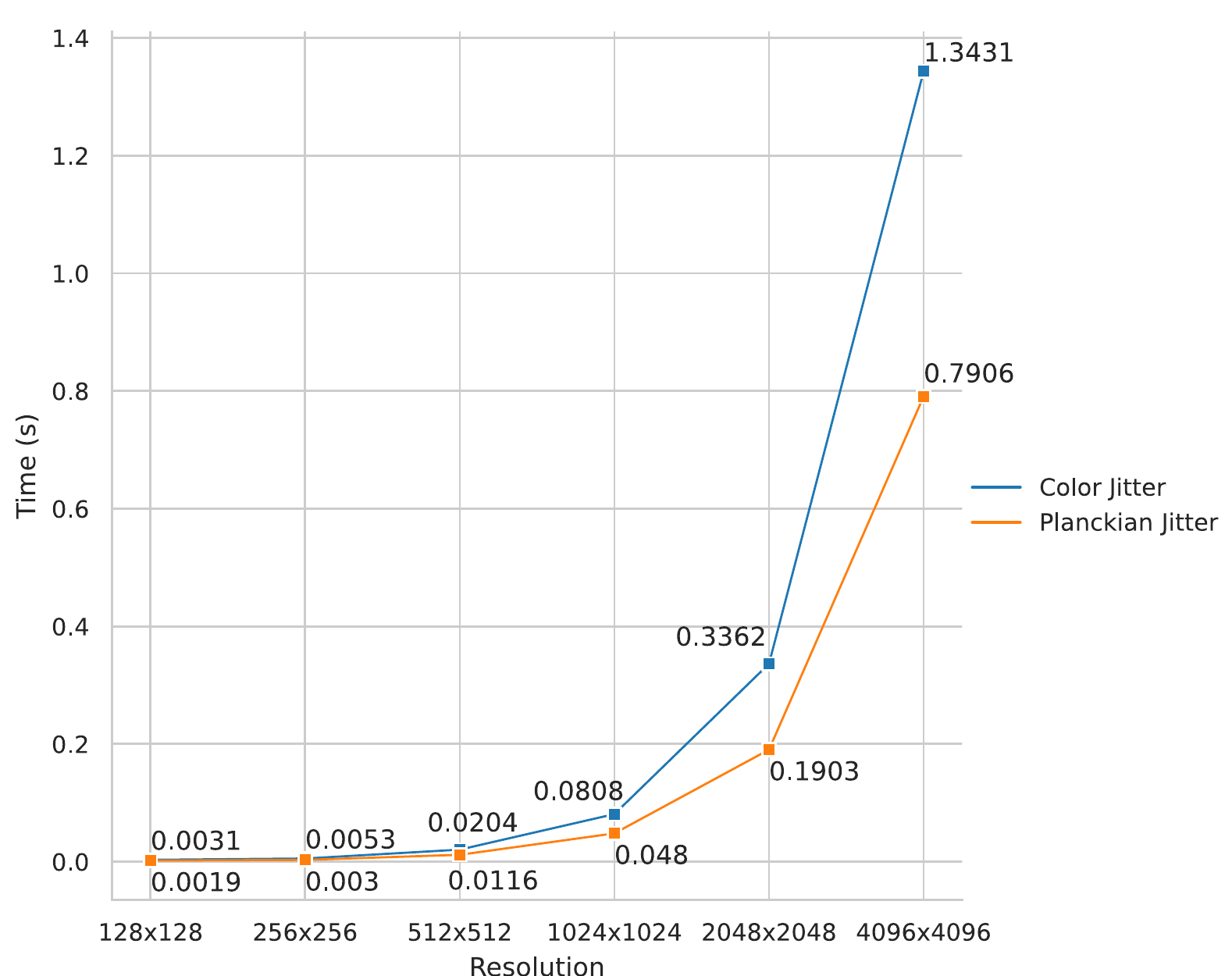}
    \caption{Comparison of execution time between the proposed \textit{Plackian Jitter} transform and the Color Jitter implementation in Pytorch Torchvision v0.9.1. For each resolution we executed both the algorithms 40 times.}
    \label{fig:time_comparison}
\end{figure}

Here we provide an analysis of execution time to assess the usability of the Planckian Jitter compared to standard Color Jitter. We executed the two algorithms: the Color Jitter image transform from Torchvision (Torch version v1.8.1 and Torchvision version v0.9.1) and our \textit{Planckian Jitter} at different image resolutions.
For each resolution we ran the code 40 times and averaged the execution time. Results are shown in Figure~\ref{fig:time_comparison}.
All augmentations were performed in CPU on an Intel i7-8700 processor. As can be seen, the proposed \textit{Planckian Jitter} is faster than  standard Color Jitter.

\subsection{Additional downstream results on Tiny-Imagenet}
\label{app:extratiny}

We also performed experiments for several other configurations of the downstream tasks with the representation trained on Tiny-ImageNet.
In Table~\ref{tab:tinyimagenet_extra} we report results for the main task and downstream task (as in section 4.4 of the main paper ImageNet, but here all images are at $64\times64$ pixel resolution.

These additional comparisons confirm the conclusions described in section 4.4 of the main paper. For all of the considered downstream tasks the application of the proposed data augmentation procedure improves the results even in comparison with other combinations of the originally used data augmentations. Moreover, the comparison with the latent space combination with the two versions of the default color jitter shows how exploiting features extracted by the model trained using the proposed Planckian Jitter augmentation enriches the expressive power of the final model.

\begin{table}[t]
\centering
\caption{Additional analysis on downstream tasks. Self-supervised training is performed on \textsc{Tiny-ImageNet} at ($64\times64$).}
\label{tab:tinyimagenet_extra}
\resizebox{\linewidth}{!}{
\begin{tabular}{lrrrrr}
\bf DATA AUGMENTATION & \multicolumn{1}{l}{\bf \textsc{Tiny-ImageNet}} & \multicolumn{1}{l}{\bf \textsc{Flowers-102}} & \multicolumn{1}{l}{\bf \textsc{Cub200}} & \multicolumn{1}{l}{\bf \textsc{VegFru}} & \multicolumn{1}{l}{\bf \textsc{T1K+}} \\\hline\\[-1.8ex]
None & 27.06\%                           & 37.65\%                         & 18.76\%                    & 24.07\%                    & 35.82\%                  \\
Default Color Jitter (CJ) & 33.12\%                           & 46.27\%                         & 19.36\%                    & 23.92\%                    & 26.01\%                  \\
Default Color Jitter w/o Grayscale (CJ-) & 31.62\%                           & 40.39\%                         & 21.90\%                    & 27.39\%                    & 32.50\%                  \\

Planckian Jitter (PJ) & 30.95\%                           & 52.35\%                         & 25.12\%                    & 28.94\%                    & 32.51\%                  \\
LSC: [CJ,CJ-]	& 39.02\%	& 58.33\%	& 26.82\%	&36.43\%	&37.20\% \\
LSC: [CJ,PJ] & 39.23\%                           & 61.57\%                         & 30.45\%                    & 39.65\%                    & 38.20\%                  \\\hline
\end{tabular}
}
\vspace{-3mm}
\end{table}

\subsection{Dependence on resolution}
\label{app:resolution_importance}
To better understand the difference of the performances reported in Tables~\ref{tab:aug_cifar_res}~and~\ref{tab:imagenet}, we perform an experiment that shows that the relative importance of texture and shape increases with increased resolution. 

As an additional experiment, we took the representations and classifiers learned at high-resolution (224x224) and investigated their sensitivity to high-frequency information in images. At inference time, we down-sample (down-sample resolution is given in Table~\ref{tab:resizing}) and then up-sample all images. In this way, we can compare the dependence on high-resolution information of different methods. Note, that here we do not retrain the classifier but use the one trained at 224x224. The results clearly show that CJ suffers more from down-sampling than PJ. For the Flowers datasets, the results of CJ at resolution 224 are better than PJ. However, when we down-sample to 64, the results change and results for PJ are already significantly better than CJ. This suggest that the texture information (important for CJ) is removed and this hurts performance. For PJ, which is more dependent on color information, down-sampling hurts results less (note PJ is also using texture, shape but to a lesser degree, so results still deteriorate for smaller resolutions). 

\begin{table}[tb]
\centering
\caption{Classification accuracy as a function of down-sampling size on Flowers. Results confirm that PJ is less sensitive to down-sampling than CJ.}
\label{tab:resizing}
\begin{tabular}{lrrrr}
\bf Method & \bf \textsc{32} & \bf \textsc{64} & \bf \textsc{128} & \bf \textsc{224}\\
\hline
CJ       &  7.74 & 49.23 & 91.43 & 91.90 \\
CJ+PJ    & 16.30 & 66.43 & 93.01 & 93.45 \\
PJ       & 23.11 & 70.13 & 89.04 & 89.22 \\
\hline
\end{tabular}
\end{table}

\subsection{Color importance in Planckian Jitter based representation}
\label{app:color_importance}
To verify that CJ uses less color information than PJ, we did a simple experiment where at inference time we changed the input images from sRGB to gray-scale images. These results are provided in Table~\ref{tab:flowers_gs}. These results clearly show that PJ is much more dependent on color than CJ. PJ has a drop of over 67.6\% whereas CJ only drops 3.2 percentage points. 

\begin{table}[tb]
\centering
\caption{Methods evaluated with color and grayscale images on the Flowers dataset.}
\label{tab:flowers_gs}
\begin{tabular}{lrr}
\bf Method & \bf \textsc{Color} & \bf \textsc{Accuracy} \\
\hline
CJ       & COLOR   &      92.73 \\
CJ       & GS      &      89.51 \\
PJ       & COLOR   &      88.97 \\
PJ       & GS      &      21.38 \\
\hline
\end{tabular}
\end{table}

\end{document}